\documentclass[letterpaper, 10 pt, conference]{ieeeconf}  

\IEEEoverridecommandlockouts                              

\usepackage{amsmath} 
\usepackage{amssymb}  
\usepackage{bm}
\usepackage{graphicx}
\usepackage{subfigure}
\usepackage{float}
\usepackage{multirow}
\usepackage{booktabs}
\usepackage{cite}
\usepackage{textcomp}
\usepackage{makecell}
\usepackage{threeparttable}
\usepackage[linesnumbered,ruled]{algorithm2e}

\providecommand{\norm}[1]{\left\lVert#1\right\rVert}

\usepackage{xcolor, soul}
\sethlcolor{white}

\makeatletter
\let\NAT@parse\undefined
\makeatother
\usepackage{hyperref}

\hypersetup{
	colorlinks,
	citecolor=blue,
}

\title{\LARGE \bf
Boosting Real-Time Driving Scene Parsing with Shared Semantics
}

\author{Zhenzhen Xiang$^{1}$, Anbo Bao$^{1}$, Jie Li$^{2}$ and Jianbo Su$^{1}$
\thanks{This work was partially financially supported by the projects of National Natural Science Foundation of China under grant 61533012 and 91748120.}
\thanks{$^{1}$Zhenzhen Xiang, Anbo Bao and Jianbo Su are with the Department of Automation, Shanghai Jiao Tong University, Shanghai 200240, China. Emails: 
	{\tt\small zzxiang.sjtu@gmail.com, \{ab\_bao, jbsu\}@sjtu.edu.cn}
}
\thanks{$^{2}$Jie Li is with SAIC Motor Corporation Limited, Shanghai 201804, China. Email:
	 {\tt\small lijie06@saicmotor.com}
}}

\begin{document}

\maketitle
\thispagestyle{empty}
\pagestyle{empty}

\begin{abstract}

Real-time scene parsing is a fundamental feature for autonomous driving vehicles with multiple cameras. \hl{In this letter we demonstrate that sharing semantics between cameras with different perspectives and overlapped views can boost the parsing performance when compared with traditional methods, which individually process the frames from each camera.} Our framework is based on a deep neural network for semantic segmentation but with two kinds of additional modules for sharing and fusing semantics. On the one hand, a semantics sharing module is designed to establish the pixel-wise mapping between the input images. Features as well as semantics are shared by the map to reduce duplicated workload which leads to more efficient computation. On the other hand, feature fusion modules are designed to combine different modal of semantic features, which leverage the information from both inputs for better accuracy. To evaluate the effectiveness of the proposed framework, we have applied our network to a dual-camera vision system for driving scene parsing. Experimental results show that our network outperforms the baseline method on the parsing accuracy with comparable computations.

\end{abstract}

\section{Introduction}

With the development of autonomous driving in recent years, scene parsing as a critical functionality of autonomous vehicles, has attracted more and more attention\cite{siam2018comparative}. Since scene parsing is a dense classification problem, it still remains a difficult task to achieve an accurate performance for real-time applications, especially for vehicles with multiple cameras and limited computational resources. 

\begin{figure}[!tb]
	\centering
	\subfigure[Configuration of our dual-camera vision system]{
		\centering
		\includegraphics[width=0.4\textwidth]{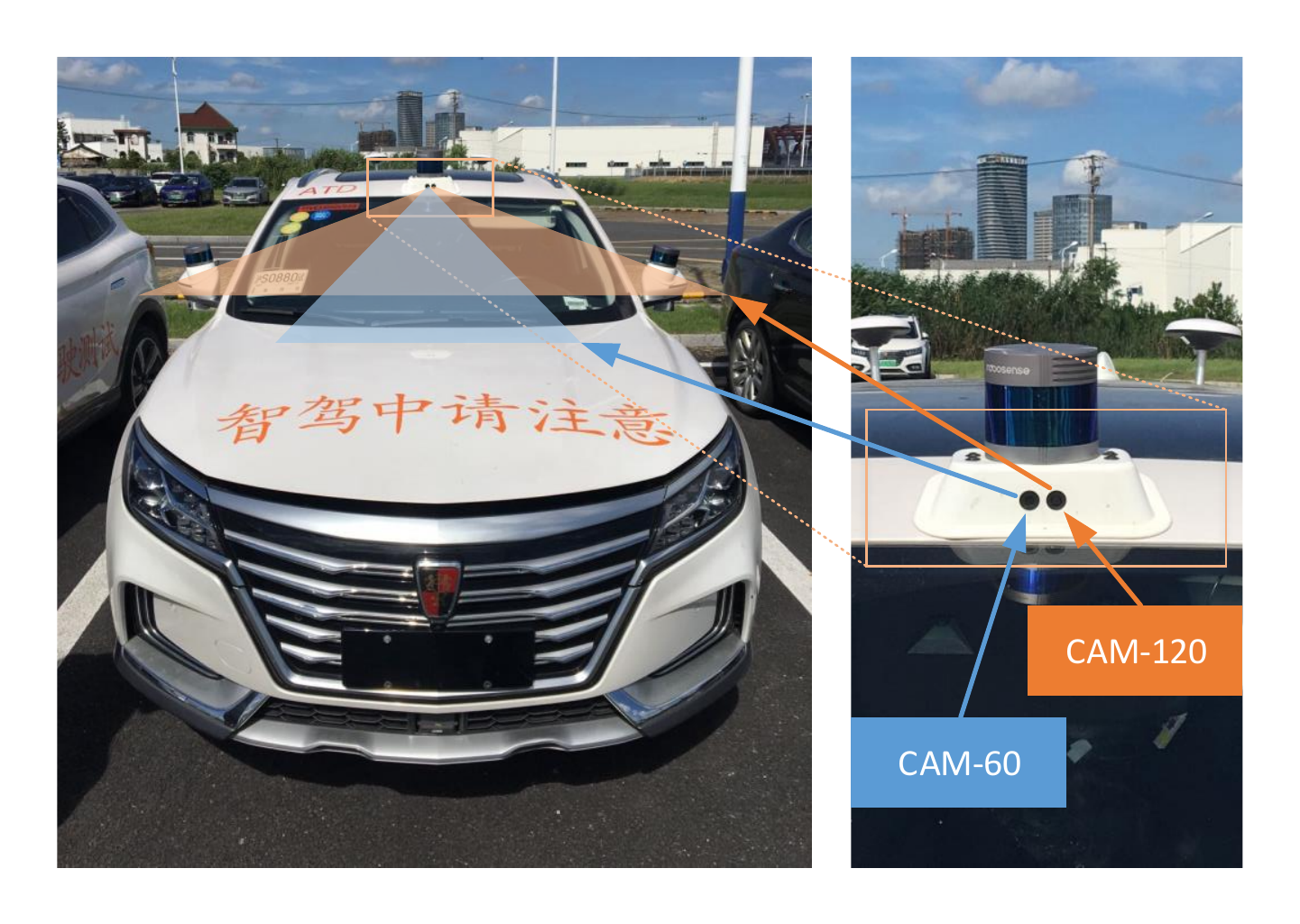}
		\label{dual_camera_config}
	}
	
	\vspace{-0.1in}
	\subfigure[Motivation of shared semantics]{
		\centering
		\includegraphics[width=0.45\textwidth]{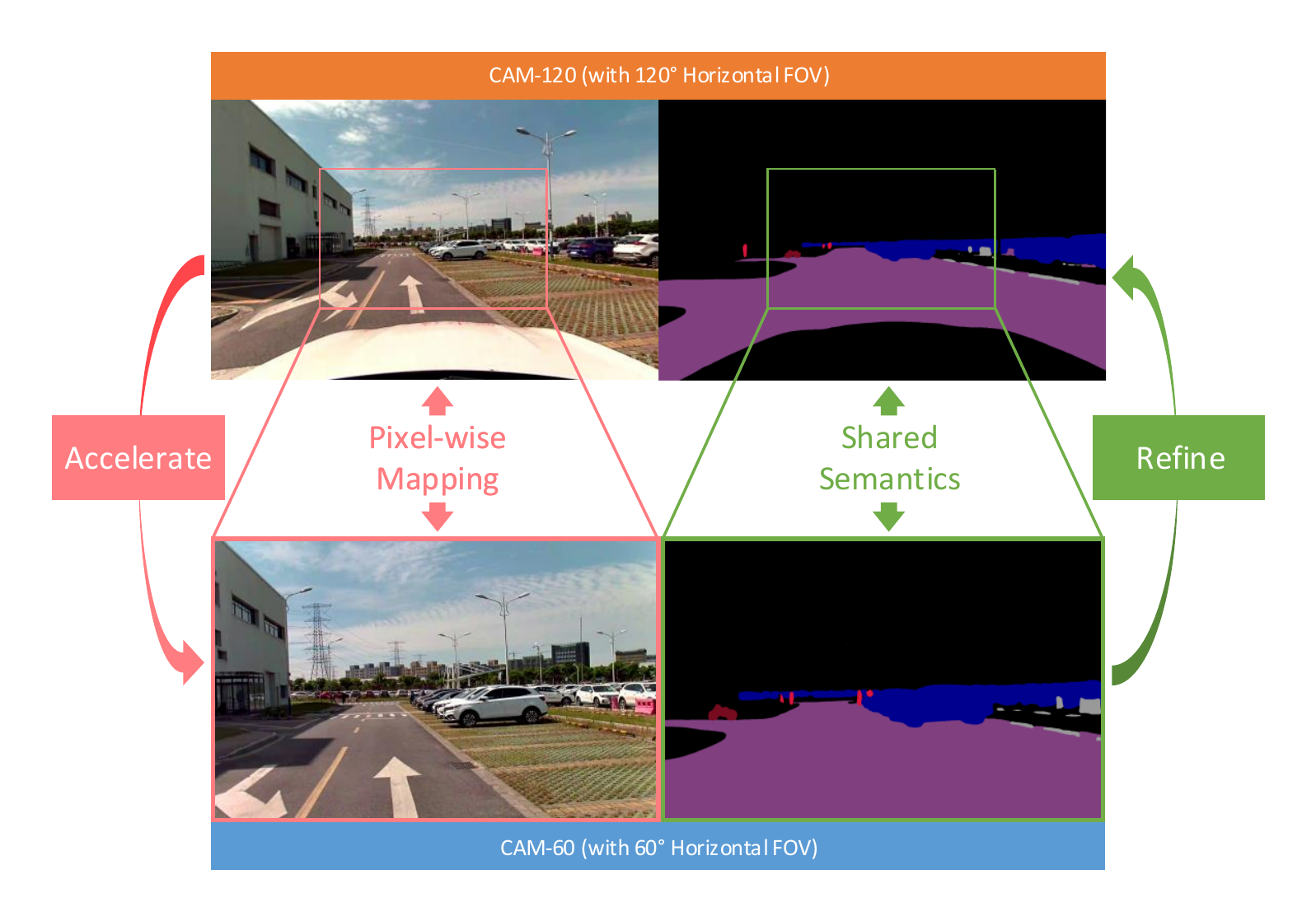}
		\label{motivation}
	}
	
	\caption{Illustration of our dual-camera system and motivation. Shared semantics based on the pixel-wise mapping accelerates the processing of CAM-60 and refines the results of CAM-120.}
	\label{fig_intro}
	\vspace{-0.15in}
\end{figure}

Taking our autonomous vehicle platform shown in Fig. \ref{dual_camera_config} as an example, a dual-camera vision system is mounted at the top of the vehicle, \hl{which is a common vision system setup adopted by modern autonomous vehicles, e.g., Tesla Autopilot\mbox{\cite{tesla}}. Compared with typical stereo cameras, these two cameras are with different field of views (FoVs), which is designed for reliable and accurate perception of objects at various distances.} In the figure CAM-60 refers to the camera with a $60^{\circ}$ horizontal FoV (HFoV) and CAM-120 stands for the camera with $120^{\circ}$ HFoV. To get scene parsing results from both cameras, traditional approaches usually process images from each camera individually, which neglects the connection inside the dual-camera system.

Since the cameras with different perspectives have overlapped perception regions as shown in Fig. \ref{motivation}, we consider to find a method to (1) build a pixel-wise mapping to share semantics between two cameras and (2) leverage the compensation of different perspectives to accelerate computation and get refined scene parsing results. More specifically, because the scenes captured by CAM-60 are almost completely contained in the image from CAM-120, processing images from CAM-60 can benefit from the information propagated from CAM-120, which leads to a more efficient computation. At the same time, because CAM-60 has a larger focal length, it has a clearer perception to the scenes far away from the vehicle. Thus CAM-120 can fuse such information to enhance its original segmentation results.

\hl{In general, when compared with traditional approaches, our method mainly boosts the scene parsing task for multi-camera systems like our configurations in the following two aspects}:

\begin{itemize}
	
	\item \hl{Reduce the computation load for cameras with narrower FoVs. Feature extraction procedure only needs to be done once in the overlapped regions among different cameras. For example, the heavy and slow feature extraction backbone for CAM-60 can be replaced with a lightweight one for extracting complementary features. The semantic information propagated from CAM-120 by a semantic sharing module provides a coarse segmentation for CAM-60, which can be further refined by fusing with its own features.}
	
	\item \hl{Improve the scene parsing quality for cameras with broader FoVs}. For example, the semantic features from CAM-60 are also back-propagated to CAM-120 with the same semantic sharing module. By appropriately fusing with the original semantics of CAM-120, those semantics located in the overlapped regions can be further enhanced with the perspective advantage from CAM-60. 
	
\end{itemize}

\section{Related Work}

\subsection{Real-Time Scene Parsing}

\hl{Deep learning based scene parsing has been extensively investigated in recent years, e.g., FCNs\mbox{\cite{long_fully_2015}}, SegNet\mbox{\cite{badrinarayanan_segnet:_2017}}.} Although the state-of-the-art semantic segmentation networks can output high-quality results \cite{zhao2017pyramid, chen2018encoder}, they are too heavy and computationally expensive to be adopted in real-time applications. Recently some lightweight semantic segmentation networks has been designed to work on-line while giving decent outputs\cite{yu2018bisenet, zhao2018icnet, howard2019searching}. However, these networks are not naturally designed for those vision systems with multiple cameras, which makes them still too memory or computationally consuming for autonomous driving applications. In our work, we aim to design an optimized architecture to reduce the redundant computation which leads to a more efficient framework.

\subsection{Semantics Sharing}
\label{rel_work_sem_sharing}

Semantics sharing seeks to find correspondences between different images which have overlapped views, e.g., the image pairs from stereo cameras or the consecutive frames in a video sequence. Semantics sharing is commonly conducted in two levels: pixel-level and feature-level.

\subsubsection{Pixel-level Sharing}

For pixel-level sharing, a pixel-wise grid map is built to warp an image from one perspective to the other. The map can be derived from the transformations in geometry space or image space. The transformation in geometry space generally uses the prior knowledge, e.g., the planar assumption for perspective transformation\cite{yin2018geonet}, or the depth estimation of the scene\cite{mayer2016a, godard2017unsupervised}. The transformation in image space usually considers the correlations around neighborhoods of a pixel\cite{dosovitskiy2015flownet}. With recent development in lightweight optical flow estimation networks\cite{hui2018liteflownet, sun2018pwc}, it is much more practical to exploit an optical estimation network in real-time applications. Xu et al. \cite{xu2018dynamic} applied different segmentation strategies to various regions of the input image, which exploited optical flow to preserve the semantics in static regions. Zhu et al. \cite{zhu2019improving} investigated the generation of future semantic segmentation labels from current manual labels by video prediction based on motion vector estimation. Yin et al. \cite{yin2018geonet} combined a rigid structure reconstructer and a non-rigid motion localizer to warp from one view to the other. 

\hl{Compared with pure geometry-based methods, the image-based methods are more robust to the errors of camera calibration and time synchronization among different cameras.} Therefore, similar to \cite{yin2018geonet}, our framework also integrates both geometry-based methods and image-based methods for sharing semantic information between two cameras at the pixel-level.

\subsubsection{Feature-level Sharing}

Feature-level sharing propagates information implicitly in the model, which is usually applied in video sequence processing. Jin et al. \cite{jin2017video} designed a network to learn predictive features in video scene parsing. Li et al. \cite{li2018low} proposed a framework with adaptive feature propagation for high-level features to reduce the latency of video semantic segmentation. Wang et al. \cite{wang2019learning} used an unsupervised method to learn feature representations for identifying correspondences across frames. Lee et al. \cite{lee2019sfnet} attempted to derive semantic correspondences by object-aware losses. Compared with pixel-level sharing, feature-level sharing is learned by an end-to-end process, and it is thus difficult to directly evaluate its performance. In addition, the feature-level sharing may rely on the training data more heavily than the pixel-level sharing used in our framework.

\subsection{Semantics Fusion}

The idea of semantics fusion for improving the segmentation outputs has been widely applied in previous works. For example, in \cite{yu2018bisenet} and \cite{zhao2018icnet}, different modals or levels of features were fused with each other to generate refined results. Li et al. \cite{li2017foveanet} hypothesized a scaled region from the original image by a perspective estimation network, which aimed to refine original segmentation results of small objects. Jiao et al. \cite{jiao2019geometry} proposed to improve and distill the semantic features with the estimated depth embeddings by geometry-aware propagation. All of the works above focus on the fusion for a single image, while Hoyer et al. \cite{hoyer2019short} demonstrated a spatial-temporal fusion method for multiple camera sequences but with non-overlapped view. In our work, we have followed the basic idea of semantics fusion and applied it to cameras with different perspectives and shared visions to enhance the overall scene parsing performance.

\section{Methodology}

In this section, we will describe the proposed method in detail. First the overview of our framework will be demonstrated. Then the ideas behind the design of each core module will be discussed. The detailed implementation information will be given at the end of the section.

\begin{figure*}[!t]
	\centering
	\includegraphics[width=0.9\textwidth]{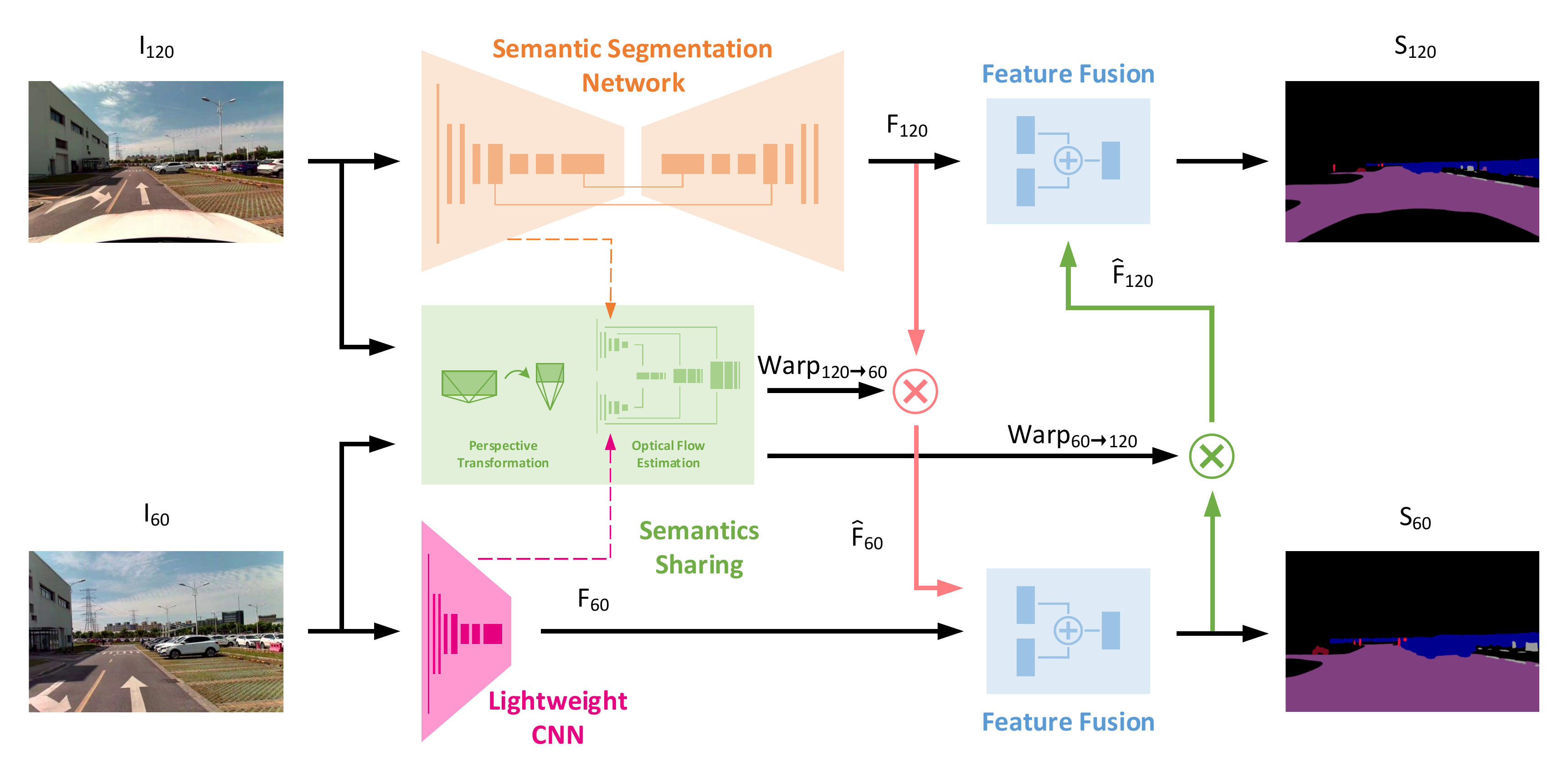}
	\vspace{-0.1in}
	\caption{The proposed scene parsing framework with shared semantics for our dual-camera vision system. The semantic segmentation network can be any real-time encoder-decoder type segmentation models. The lightweight CNN extracts features fast for CAM-60. The semantics sharing module consists of perspective transformation and optical flow estimation, which provides maps for warping between CAM-120 and CAM-60. Note that the dashed orange and magenta arrows means the optical flow estimation reused the features from the semantic segmentation network and the lightweight CNN. Please refer to Sec.~\ref{imple_structure} for more details. The feature fusion modules merge the shared semantics for final results.}
	\label{framework}
	\vspace{-0.2in}
\end{figure*}

\subsection{Framework Overview} 

The proposed framework is illustrated in Fig. \ref{framework}. The final goal is to output the scene parsing results for each input image from both CAM-120 and CAM-60.

From the view of structure, our framework can be divided into two branches. Unlike traditional designs with exactly the same pipeline for both branches, the input image from CAM-60 passes a much more lightweight convolutional neural network (CNN) compared with a complete semantic segmentation network in the branch of CAM-120. The sharing and fusion of information between two branches are realized with a semantics sharing module and two feature fusion modules, respectively.

From the view of functionalities, four kinds of modules in our framework play different roles. \hl{The semantic segmentation network provides high-level semantic features for sharing. The lightweight CNN recovers the detailed and complementary features for refined parsing results, similar to the architecture of ICNet\mbox{\cite{zhao2018icnet}}}. The semantic sharing module establishes a bridge for bi-directional feature propagation from CAM-120 to CAM-60 and vise versa. The feature fusion modules merge shared semantics for each branch to achieve better parsing results.

Since the semantic segmentation network is a full-function network which can output scene parsing results by itself, it can be easily replaced with any modern networks designed for real-time scene parsing. For the lightweight CNN, it can also be designed as a sequential of several convolutional layers or sharing the structure with the feature extraction backbone in the semantic segmentation network. The implementation details of these two parts will be described in Sec.~\ref{method_imple_det}. In the following we will focus on the details of the semantic sharing module and the feature fusion module.

\begin{figure*}[!t]
	\centering
	\includegraphics[width=0.9\textwidth]{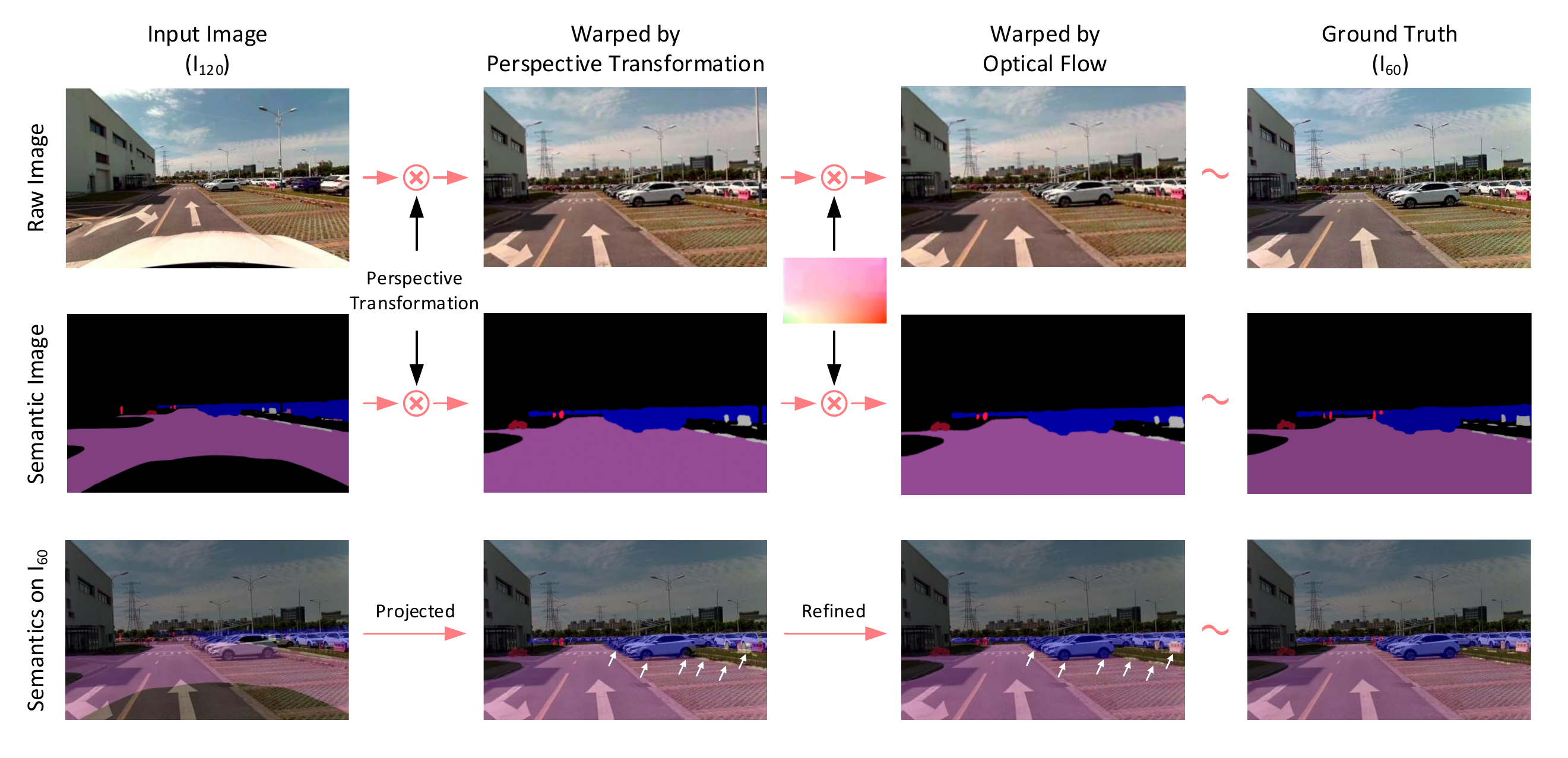}
	\vspace{-0.15in}
	\caption{The pipeline of semantics sharing. The first and second row show the warping progress for the raw and semantic image of CAM-120, respectively. The third row demonstrates the propagated semantics from CAM-120 to CAM-60, where the white arrows point out the improvements after refinement.}
	\label{semantic_sharing}
	\vspace{-0.15in}
\end{figure*}

\subsection{Semantics Sharing Module}

The task of the semantics sharing module is to remap the semantic features between two branches. Through such a bridge, the semantic features from branch CAM-120 can be propagated to branch CAM-60 to speed-up its processing, and then the results of CAM-60 are transfered back to refine the outputs of CAM-120, which forms a closed loop.

\hl{Although the pure geometry-based method (e.g., the depth estimation based warping) and feature-level propagation can also be used for sharing semantics, as concluded in Sec.~\mbox{\ref{rel_work_sem_sharing}}, the results of pixel-level sharing methods are more robust and explicit,} and thus we have proposed a two-stage image warping method to build the semantics sharing module as shown in Fig.~\ref{semantic_sharing}.

\subsubsection{Stage I: Geometry-based Warping}

In the first stage, the input image from CAM-120 is warped by the perspective transformation. The homography matrix used in the transformation can be derived from the intrinsic and extrinsic parameters of the dual-camera system \cite{szeliski2010computer}:
\begin{equation}\label{eq_homography}
\begin{aligned}
H_{120\rightarrow60} = K_{60}RK_{120}^{-1},
\end{aligned}
\end{equation}
where $ H_{120\rightarrow60} $ is the homography matrix for mapping from CAM-120 to CAM-60, $ K_{60} $ and $ K_{120} $ are their camera matrices. $ R $ is the rotation matrix from CAM-120 to CAM-60. Due to the limitation of perspective transformation, those objects closed to the camera will be distorted after the transformation. Thus the warped image from CAM-120 still needs to be adjusted to accurately match the ground truth image from CAM-60.

\subsubsection{Stage II: Image-based Warping}

In the second stage, the warped image from CAM-120 is further warped by the optical flow to compensate the distortion effects. The core process of this stage is the precise estimation of optical flow between the input image pairs. It should be noticed that because the pose variation between two cameras is very small and the input image pairs are correctly synchronized, the scene can be considered as static and the occlusion effect is negligible. Therefore the movement of pixels is not that large and the artifacts of the warped image also can be ignored comparing with the situation of video scene parsing.

\begin{figure}[!t]
	\centering
	\subfigure[Basic type]{
		\centering
		\includegraphics[width=0.3\textwidth]{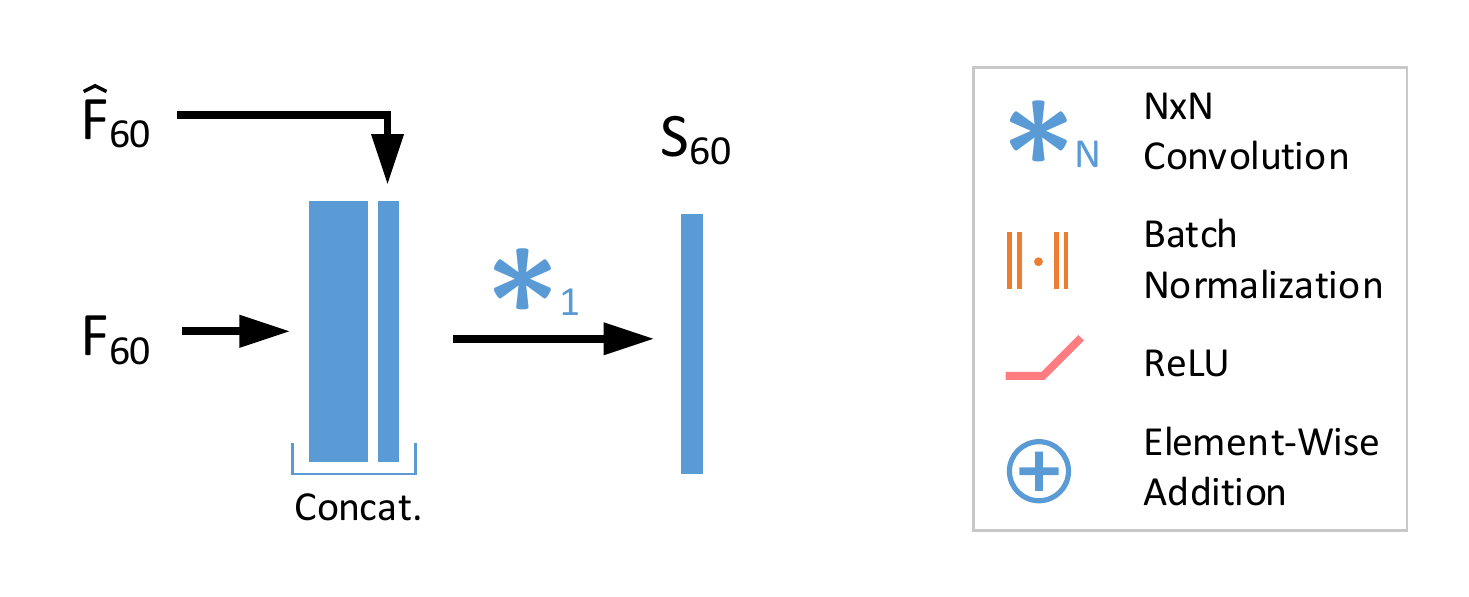}
		\label{ffm_basic}
	}
	
	\vspace{-0.3in}
	\subfigure[Residual type]{
		\centering
		\includegraphics[width=0.3\textwidth]{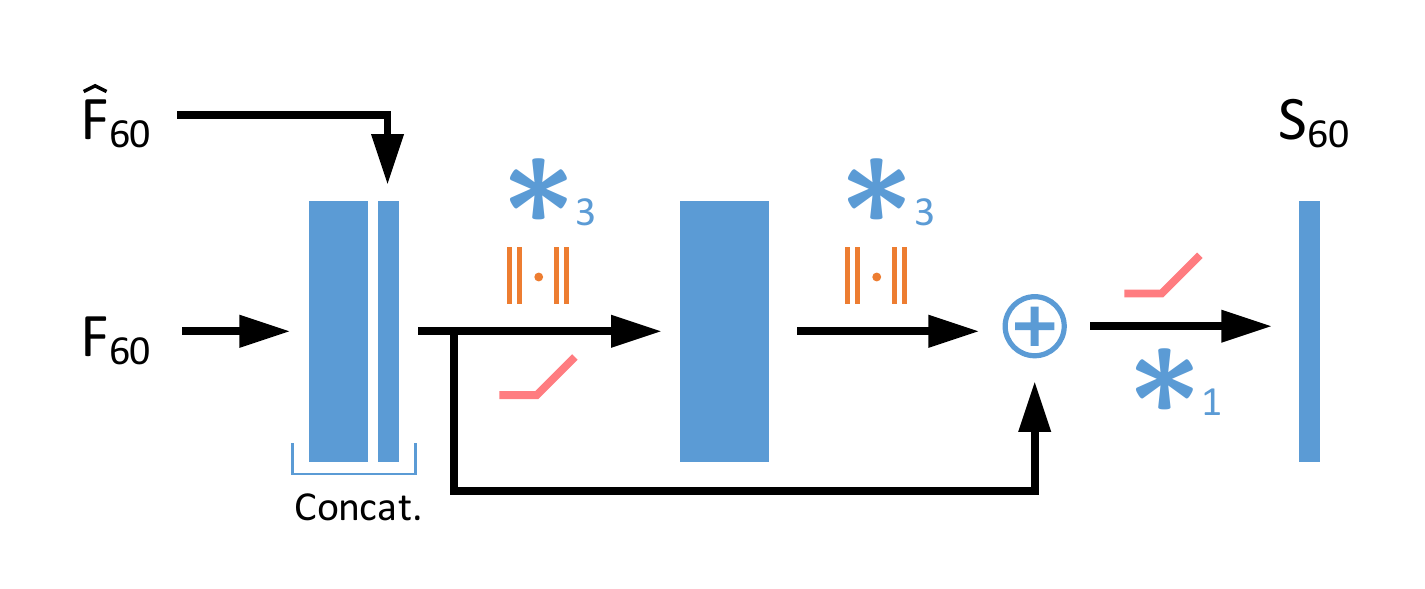}
		\label{ffm_residual}
	}
	
	\vspace{-0.3in}
	\subfigure[Bottleneck type]{
		\centering
		\includegraphics[width=0.35\textwidth]{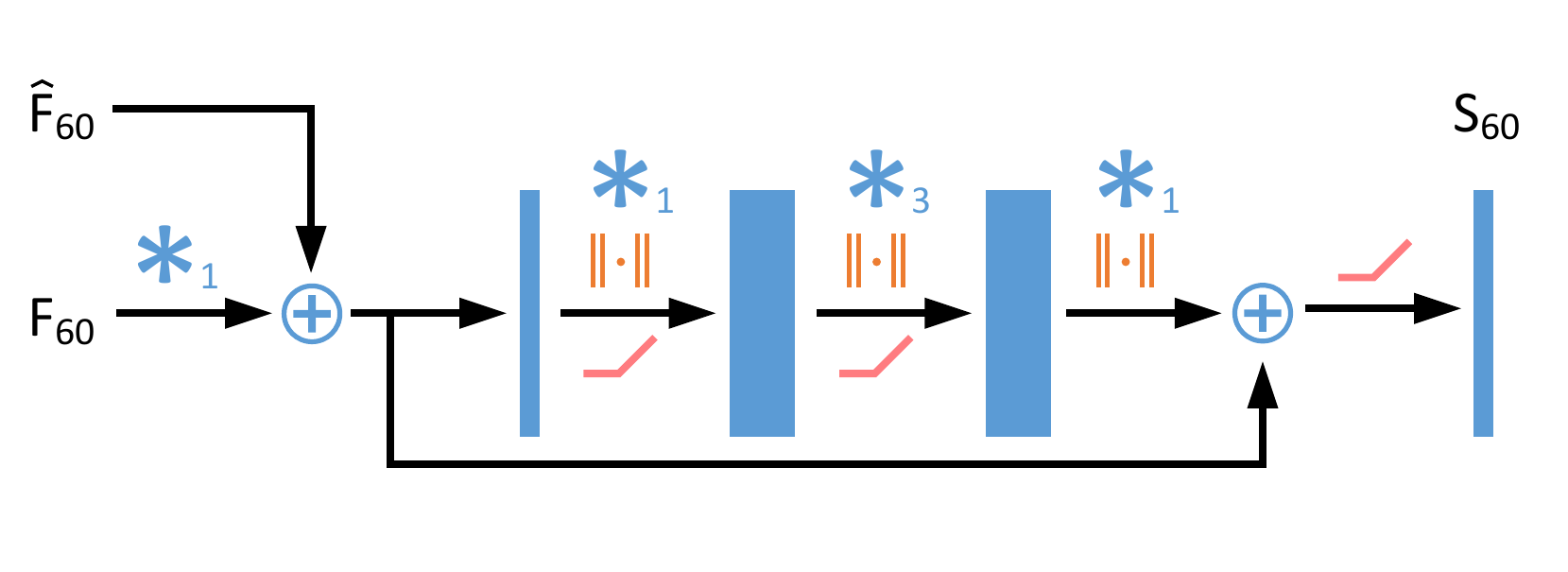}
		\label{ffm_bottleneck}
	}
	\vspace{-0.1in}
	\caption{Three optional types of feature fusion modules in our framework.}
	\vspace{-0.15in}
	\label{ffm}
\end{figure}

\subsection{Feature Fusion Module}

The feature fusion modules are used to generate the segmentation results of CAM-60 as well as to refine the results of CAM-120. As shown in Fig.~\ref{ffm}, we have implemented and evaluated three different types of the feature fusion modules to compare with the direct output of the semantic segmentation network. In Sec.~\ref{exp_compare_ffm} we will show the ablation analysis of these blocks which depicts that even integrating the simple basic block can boost the parsing outputs to some extent.

In the following we will take the feature fusion module in the CAM-60 branch as an example to describe their structures.

\subsubsection{Basic Type}

The basic type of feature fusion module only concatenates the input feature maps and output the semantic feature maps after an $ 1\times1~conv $.

\subsubsection{Residual Type}

Since the effectiveness of residual block has been widely proved in previous works, we also apply it to our framework. The inputs are first concatenated and then passed through a standard residual block with $ 3\times3~conv $ layers. Finally the output is processed by an $ 1\times1~conv $ for classification.

\subsubsection{Bottleneck Type}

Considering to decrease the computation and the amount of parameters in our framework, we also evaluate a bottleneck type of feature fusion module. The inputs are first converted to the same channels with an $ 1\times1~conv $, then they are passed through a bottleneck with an expansion, followed by an rectified linear unit (ReLU) for the output.

\subsection{Implementation Details}
\label{method_imple_det}

\subsubsection{Structure}
\label{imple_structure}

Taking the implementation of the semantics sharing module into account, we have developed two types of structures for our framework: a) loosely-coupled structure and b) tightly-coupled structure. For the loosely-coupled structure, we simply exploit a complete optical flow network following the perspective transformation, which can achieve the best estimation performance.

\begin{figure}[!t]
	\centering
	\includegraphics[width=0.4\textwidth]{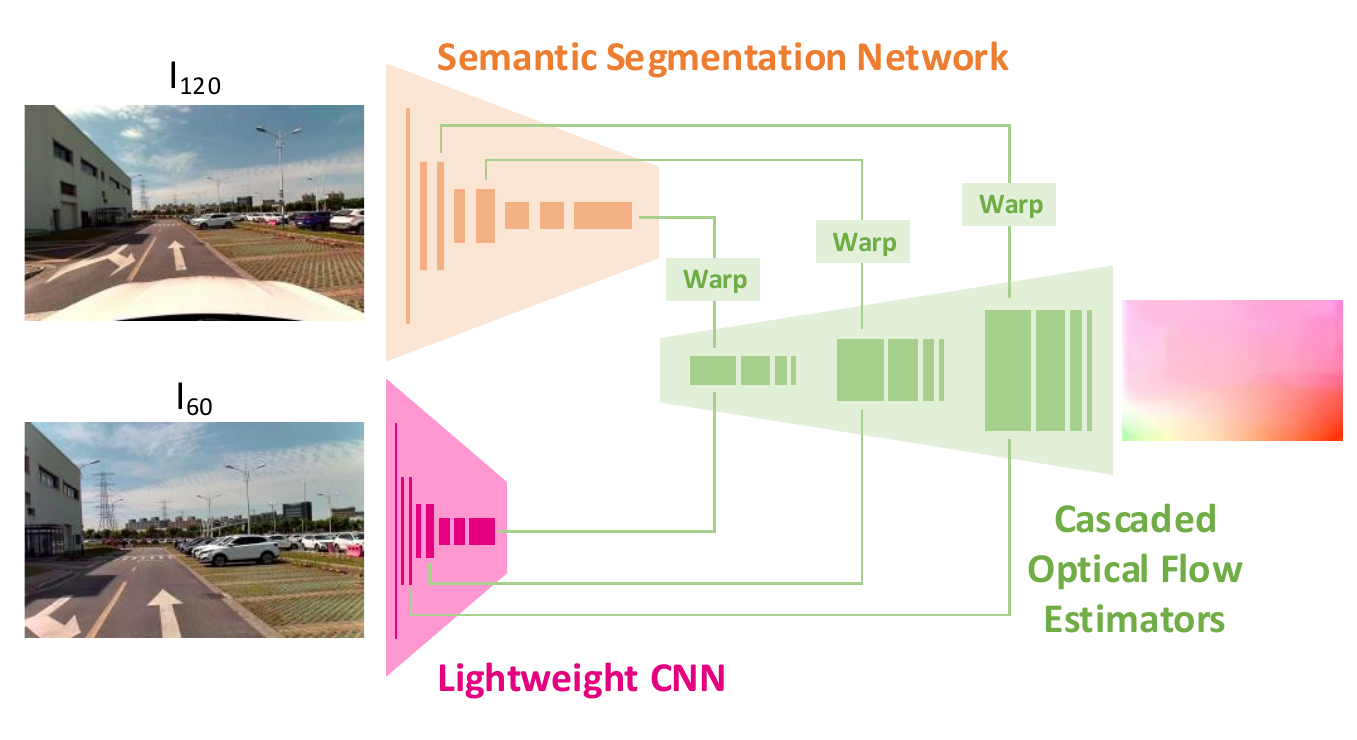}
	\caption{Tightly-coupled structure of optical flow estimation network.}
	\label{optical_flow_tight}
	\vspace{-0.2in}
\end{figure}

However, because the optical flow network also has its own feature extraction modules, it is possibly duplicable to those in the semantic segmentation network. Therefore, in the tightly-coupled structure shown in Fig.~\ref{optical_flow_tight}, we remove the feature extraction part of the optical flow network and reused the feature maps from the semantic segmentation network. With such adjustments made, the whole model becomes more compact and the computation load can be further cut down.

\subsubsection{Semantic Segmentation Network}

We exploit a real-time semantic segmentation network based on MobileNetV3-large \cite{howard2019searching} to get the initial semantic features of CAM-120. To train the semantic segmentation network, we apply the common cross entropy loss to supervise the training progress.

\subsubsection{Lightweight CNN}

The implementation of the lightweight CNN is based on the structure of the framework. For the loosely-coupled structure, we share the structure with the semantic segmentation network and output a feature pyramid with 1/8 size of the original resolution for later fusion with the results from CAM-120. For the tightly-coupled structure, we reuse the feature extraction part of the optical flow network and adjust it to output a feature pyramid with exactly the same size from 1/4, 1/8 to 1/16 as those from the semantic segmentation network.

The lightweight CNN also shares the weights with the semantic segmentation network in the loosely-coupled structure. In the tightly-coupled structure, it is trained together with the optical flow network.

\subsubsection{Semantics Sharing Module}

The main part of the semantics sharing module is an optical flow estimation network. We use a PWC-Net \cite{sun2018pwc} to provide grid maps for warping feature maps. It should be noticed that the original feature pyramid given by PWC-Net is not the same as the MobileNetV3-large. Thus for the tightly-coupled structure, we need to modify the channels of the output feature maps to match those in MobileNetV3-large accordingly.

The training losses of the optical flow network in our cases consist of three different types: a) supervised loss, b) unsupervised loss and c) semantic loss. The supervised loss is applied when the ground-truth flow is available with some synthetic datasets. It is defined as the average end-point error (AEPE):
\begin{equation}\label{eq_loss_sup}
\begin{aligned}
\mathcal{L}_{sup} = \frac{1}{N}\sum_{i}^{N}\norm{\mathbf{w}(p_i)-\mathbf{\hat{w}}(p_i)}_2,
\end{aligned}
\end{equation}
where $p$ is the pixel index and N is the total number of pixels in the flow image. $\mathbf{w}$ and $\mathbf{\hat{w}}$ are the ground-truth and the estimated flow, respectively.

The unsupervised loss is mainly for training on those datasets without the ground-truth flow. We choose three most commonly used losses for unsupervised learning:
\begin{equation}\label{eq_loss_unsup}
\begin{aligned}
\mathcal{L}_{unsup} = w_1\mathcal{L}_{1} + w_2\mathcal{L}_{SSIM} + w_3\mathcal{L}_{smooth}.
\end{aligned}
\end{equation}
Here the first term is defined as the $L_1$ norm of the pixel intensity difference between the ground-truth image $\mathbf{I}$ and the flow-warped image $\mathbf{\hat{I}}$:
\begin{equation}\label{eq_loss_l1}
\begin{aligned}
\mathcal{L}_{1} = \frac{1}{N}\sum_{i}^{N}\norm{\mathbf{I}(p_i)-\mathbf{\hat{I}}(p_i)}_1.
\end{aligned}
\end{equation}
The second term is the SSIM \cite{wang2004image} loss of the ground-truth image and the flow-warped image. The third term is the smoothness loss \cite{heise2013pm} of the estimated flow. The weights of these three losses $w_1$, $w_2$ and $w_3$ are set to 0.1, 1.0 and 1.0, respectively.

The semantic loss is for the dataset with semantic labels. It can be regarded as a supervision for flow at the boundaries of each semantic class. Here we also applied the cross entropy loss to supervise the fine-tuning of the optical flow network.

\subsubsection{Feature Fusion Module}

For the basic and residual type of feature fusion modules, they are applied to both branches without modification. However, since the bottleneck type has an element-wise addition unit, we will additionally need an $ 1\times1~conv$ to reshape $ F_{60} $ to the same size as $ \hat{F}_{60} $ in the CAM-60 branch, as shown in Fig.~\ref{ffm_bottleneck}.

\section{Experiments}

\subsection{Dataset and Evaluation Metrics}

\begin{figure}[!t]
	\centering
	\vspace{5pt}
	\includegraphics[width=0.38\textwidth]{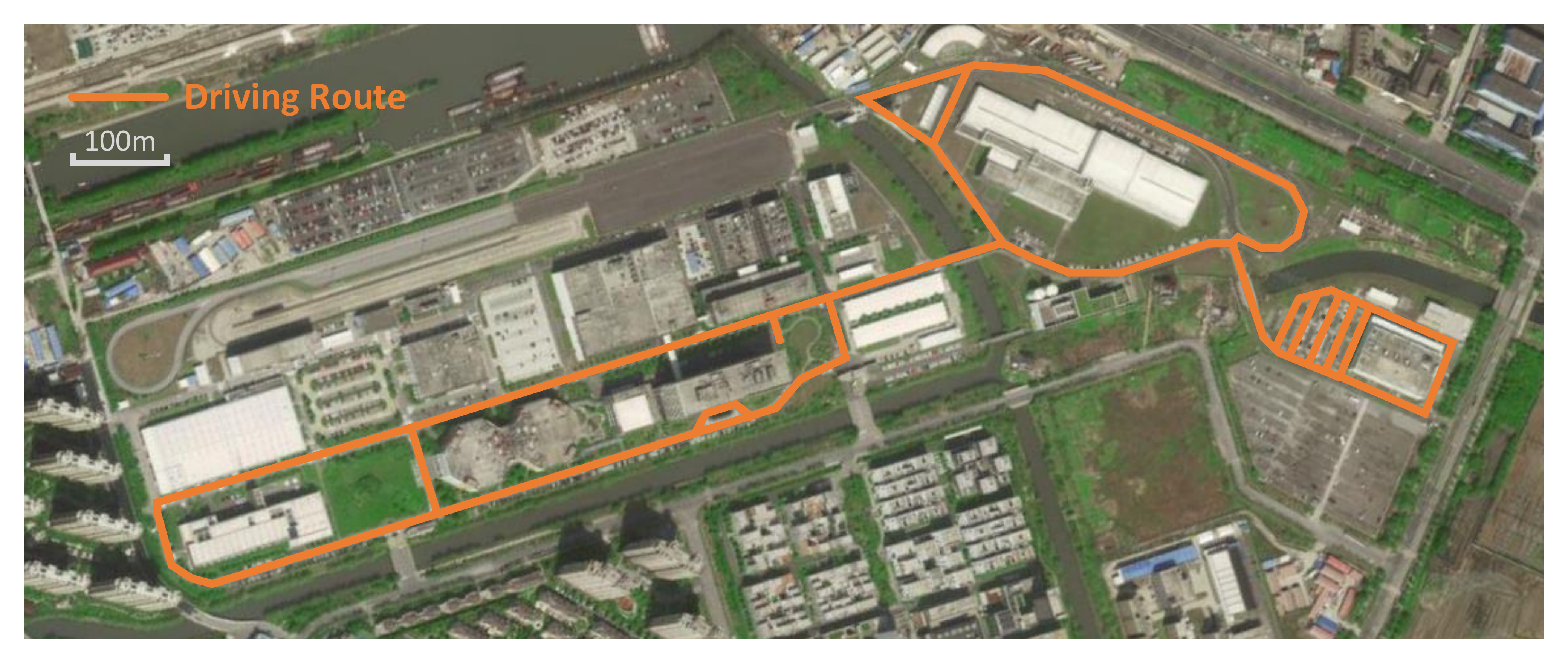}
	\caption{Driving route for the video data collection of the dataset.}
	\label{dataset}
\end{figure}

Since we have not found any public dataset with configurations as our applications, we built our own dataset with a dual-camera system on an autonomous vehicle. The videos were captured by a Sekonix SF3324 (CAM-120) and a Sekonix SF3325 (CAM-60) with an NVIDIA DRIVE AGX platform. The video sequences were collected inside the SAIC Motor Park and the driving route is shown in Fig.~\ref{dataset}. 

\begin{table}[!t]
	\renewcommand{\arraystretch}{1.3}
	\caption{Statistics of Our Dataset}
	\label{table_dataset}
	\centering
	\resizebox{0.48\textwidth}{!}{
	\begin{tabular}{|c|c|c|c|c|c|}
		\hline
		\makecell*[c]{\textbf{Length} \\ \textbf{(km)}} & \makecell*[c]{\textbf{Duration} \\ \textbf{(h)}} & \makecell*[c]{\textbf{Max. Speed} \\ \textbf{(km/h)}} & \makecell*[c]{\textbf{Image} \\ \textbf{Pairs}} & \makecell*[c]{\textbf{Image} \\ \textbf{Size}} & \makecell*[c]{\textbf{Semantics} \\ \textbf{Classes}} \\
		\hline
		$\sim$10 & 1.28 & 30 & 4593 & 1920$ \times $1208 & 6 \\
		\hline
	\end{tabular}}
	\vspace{-0.15in}
\end{table}

The statistics of the dataset is listed in Table~\ref{table_dataset}. The videos from each camera are synchronized by the hardware. We automatically extract images from the videos at the rate of one frame per second. Then we select about 1000 image pairs to manually label six classes of semantics: \textit{background (BG), road, person, car, barrier} and \textit{cycle}. We use these images to train a PSPNet \cite{zhao2017pyramid} to automatically label the other images as the ground truth for later training the semantic segmentation network based on MobileNetV3-large for real-time parsing.

In addition to the semantic labels, we also need ground-truth optical flow to train the PWC-Net, which is difficult to obtain. So we turned to synthesize a warped image by a random perspective transformation from an input image. \hl{Specifically, the focal length of CAM-120 in its camera matrix is multiplied by a random factor between 0.95 and 1.05; then the image is further randomly translated by $\pm$10 pixels and rotated by $\pm$5$^\circ$. The flows generated along with the transforming process is used as the ground truth.} In Sec.~\ref{exp_compare_traing_schedule} we will show that after training on this dataset, the performance of PWC-Net on the original dataset will be improved.

The performance of our network is evaluated with the mean intersection over union (mIoU) metric for semantic segmentation and average end-point error (AEPE) for optical flow estimation.

\subsection{Training Procedure}
We follow a multi-stage training procedure to train each component of our framework:

\begin{figure*}[!thb]
	\centering
	\vspace{0.15in}
	\includegraphics[width=0.86\textwidth]{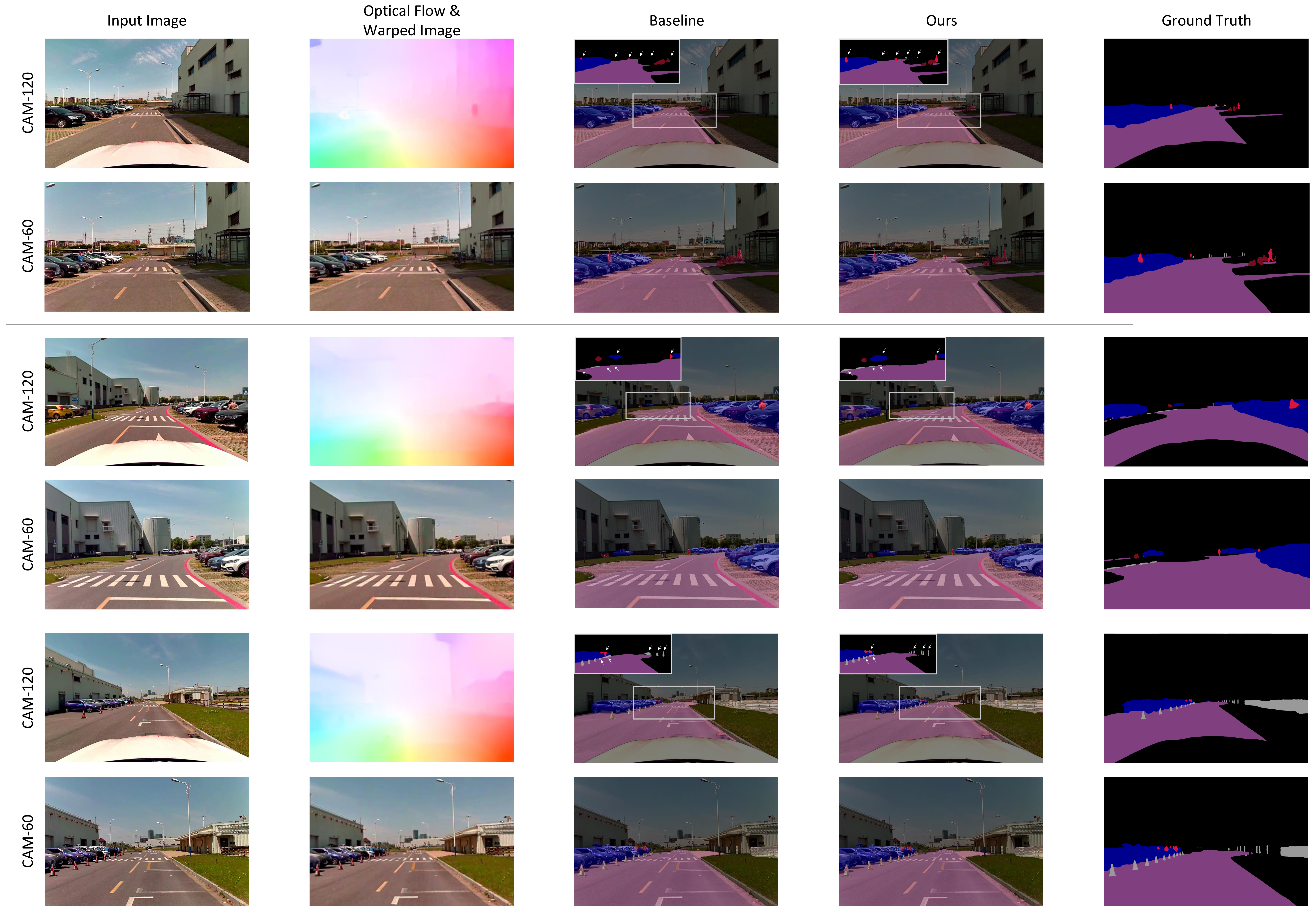}
	\vspace{-0.1in}
	\caption{Visualization of scene parsing results for the baseline and our framework. In the third and fourth column for CAM-120, the main difference between our results and the baseline are pointed out with white arrows.}
	\label{result_pic}
\end{figure*}

\begin{table*}[!tbp]
\renewcommand{\arraystretch}{1.3}
\caption{Comparisons among the baseline and our method with loosely-coupled and tightly-coupled structures.}
\label{table_main_result}
\centering
\resizebox{0.85\textwidth}{!}{
\begin{tabular}{|c|c|c|c|c|c|c|c|c|c|c|c|c|c|}
	\hline
	\multirow{2}{*}{\textbf{Network}} & \multirow{2}{*}{\textbf{CAM}} &  \multicolumn{8}{c|}{\textbf{Mean IoU of Semantic Segmentation ($\%$)}}  &  \multicolumn{2}{c|}{\textbf{Model}} & \multicolumn{2}{c|}{\textbf{\color{black}Runtime \color{black}(fps)}} \\ 
	\cline{3-14}
	& & BG & Road & Person & Car & Barrier & Cycle & Avg. & Total & Params & FLOPS & \color{black}Full Res. & \color{black}Half Res. \\
	\hline
	\multirow{3}{*}{Baseline} & 60 & 99.4 & 99.1 & \textbf{78.6} & \textbf{95.6} & 73.4 & 70.4 & 86.1 & \multirow{2}{*}{85.9} & \multirow{3}{*}{\textbf{2.81M}} & \multirow{3}{*}{34.34G} & \multirow{3}{*}{\color{black}29.8} & \multirow{3}{*}{\color{black}\textbf{103.2}}\\
	\cline{2-9}
	& 120 & 99.5 & 98.6 & 74.1 & 95.2 & 78.0 & 68.8 & 85.7 & & & & & \\
	\cline{2-10}
	& \color{black}120-OL & \color{black}99.3 & \color{black}98.8 & \color{black}72.6 & \color{black}94.3 & \color{black}72.5 & \color{black}70.1 & \color{black}84.6 & - & & & & \\
	\hline
	\multirow{3}{*}{\makecell{Loosely-\\coupled}} & 60 & \textbf{99.5} & \textbf{99.2} & 76.8 & 95.5 & \textbf{75.4} & \textbf{71.3} & \textbf{86.3} & \multirow{2}{*}{\textbf{87.0}} & \multirow{3}{*}{6.57M} & \multirow{3}{*}{39.28G} & \multirow{3}{*}{\color{black}19.6} & \multirow{3}{*}{\color{black}47.7}\\
	\cline{2-9}
	& 120 & \textbf{99.6} & \textbf{98.8} & \textbf{76.4} & \textbf{95.7} & \textbf{80.3} & \textbf{74.7} & \textbf{87.6} & & & & & \\
	\cline{2-10}
	& \color{black}120-OL & \textbf{\color{black}99.4} & \textbf{\color{black}99.0} & \textbf{\color{black}76.3} & \textbf{\color{black}95.2} & \textbf{\color{black}77.1} & \textbf{\color{black}75.7} & \textbf{\color{black}87.1} & - & & & & \\
	\hline
	\multirow{3}{*}{\makecell{Tightly-\\coupled}} & 60 & 99.4 & 98.8 & 75.8 & 94.6 & 74.6 & 69.8 & 85.5 & \multirow{2}{*}{86.3} & \multirow{3}{*}{4.50M} & \multirow{3}{*}{\textbf{33.76G}} & \multirow{3}{*}{\color{black}\textbf{31.0}} & \multirow{3}{*}{\color{black}78.1}\\
	\cline{2-9}
	& 120 & \textbf{99.6} & 98.7 & 75.2 & 95.6 & 80.1 & 72.6 & 87.0 & & & & & \\
	\cline{2-10}
	& \color{black}120-OL & \textbf{\color{black}99.4} & \textbf{\color{black}99.0} & \color{black}76.1 & \color{black}95.1 & \color{black}76.6 & \color{black}74.7 & \color{black}86.8 & - & & & & \\
	\hline
\end{tabular}}
	\vspace{-0.15in}
\end{table*}

\begin{itemize}
	
	\item Semantic segmentation network: We trained the MobileNetV3-large with a segmentation head for 160K iterations using a mini-batch size of 16. The initial learning rate was set to 0.015 and followed a `poly' policy with power 0.9.
	
	\item Optical flow estimation network: For the loosely-coupled structure, the PWC-Net was trained separately to the segmentation network. It was first trained on the \textit{FlyingChairs} dataset with the same settings as \cite{sun2018pwc}. Then we further trained it on our synthetic flow dataset for 300K iterations using a mini-batch size of 8. The initial learning rate was 0.0005 and was scaled by 0.5 at 100K, 200K, 250K. Finally the model was fine-tuned on the real data with unsupervised losses and semantic loss sequentially. For the tightly-coupled structure, only one of the feature extraction parts and the optical flow estimation part were needed to be trained. The training settings remained the same as the loosely-coupled structure.
	
	\item Feature fusion modules: The feature fusion modules were trained with the whole network with the fixed weights of MobileNetV3-large and PWC-Net. The feature fusion module in CAM-60 branch was first trained for 60K iterations with a mini-batch size of 4. The initial learning rate was set to 0.001. The other training settings were the same as MobileNetV3-large. The feature fusion module in CAM-120 was also trained in the same way.
	
	\item Fine-tuning: The whole network was finally fine-tuned together for 120K iterations with the same settings as training MobileNetV3-large. In order to keep a steady performance of optical flow estimation, the weights of PWC-Net were fixed in the final fine-tuning.
	
\end{itemize}

We use PyTorch to implement our network. The network is trained and tested on two NVIDIA Tesla V100 GPUs. \hl{Our code and trained models have been made publicly available at:} \url{https://github.com/zhenzhenxiang/SemanticsSharing}.

\subsection{Main Results}

We have chosen the MobileNetV3-large based segmentation network \cite{howard2019searching} as our baseline, which is also used in the CAM-120 branch of our framework. As shown in Table~\ref{table_main_result}, \hl{we have compared the semantic segmentation performance, model statistics and runtime of the baseline and our network with loosely-coupled and tightly-coupled structures.}

\subsubsection{Semantic Segmentation}

For the CAM-60 branch, the segmentation results show that our loosely-couped structure has slightly over-performed than the baseline in general, although there is only a lightweight CNN in this branch. Besides, the performance of our tightly-coupled structure is also very close to the baseline with more reused intermediate features.

For the CAM-120 branch, both of our loosely-coupled and tightly-coupled structure have an obvious improvement comparing with the baseline, especially for the loosely-coupled with the class of \textit{Person} (+2.3\%), \textit{Barrier} (+1.7\%) and \textit{Cycle} (+5.9\%). \hl{In addition, the mIoU results in the central view of CAM-120 which overlaps with CAM-60 (c.f. ``120-OL'') are also provided for further comparisons. The results show significant improvements on \textit{Person} (+3.7\%), \textit{Barrier} (+4.6\%) and \textit{Cycle} (+5.6\%).} This reflects the effectiveness of our semantics sharing module and feature fusion module which propagate and fuse the semantic information from CAM-60 to CAM-120. The sharing of such information compensates and improves the features of those small objects in the view of CAM-120. As shown in Fig.~\ref{result_pic}, our network successfully recovers the missing small objects that are far from the vehicle (c.f. the first and second group of image pairs), and has a more accurate classification at the boundary of small objects (c.f. the third group of image pairs).

\subsubsection{Model}

Our loosely-coupled model has 2.3$\times$ more parameters than the baseline, which can be reduced to 1.6$\times$ with tightly-coupled structure. The computation is evaluated by input images with 1920$\times$1208 resolution for the MobileNetV3-large and 768$\times$483 for PWC-Net. The results show that our loosely-coupled model has comparable computation with the baseline, while the tightly-coupled model needs even less computation resources.

\subsubsection{Runtime}

\hl{The results are the average runtime of 300 inferences for each model. As shown in Table~\mbox{\ref{table_main_result}}, the baseline and our models are all capable for real-time applications, especially when the size of input image is a half of its original resolution, i.e., 960$\times$604. When deployed to embedded devices, the models can be further optimized which will lead to a much higher frame rate.
}

\subsection{Ablation Study}

\subsubsection{Loosely-Coupled vs. Tightly-Coupled}

The influence of loosely-coupled and tightly-coupled structure for optical flow estimation was evaluated. The results on \textit{Data\_Sim} (the hypothesized image pairs) and \textit{Data\_Real} (the real image pairs) are listed in Table~\ref{table_net_structure}. We can find that the tightly-coupled structure has larger AEPE and unsupervised loss in both datasets. This is mainly because that we have fixed the weights of reused feature extraction part from MobileNetV3-large and trained the rest part of PWC-Net. From the comparisons of semantic segmentation in Table~\ref{table_main_result} for both structures, we can find that such inaccuracy of flow estimation will only remains slight effect on the segmentation results after fine-tuning the whole network.

\subsubsection{Optical Flow Estimation with different training schedules}
\label{exp_compare_traing_schedule}

Since the performance of optical flow estimation can be influenced by the training schedules on different datasets \cite{sun2018pwc}, we also evaluated the effectiveness of the synthetic \textit{Data\_Sim} dataset. Table~\ref{table_train_schedule} shows the comparisons of three different types of training schedules. It suggests that the training on \textit{Data\_Sim} has positive effects on the original network trained with the \textit{FlyingChairs} \cite{dosovitskiy2015flownet} dataset and improves its performance on the final \textit{Data\_Real} dataset.

\subsubsection{Semantic Sharing w/ or w/o Warping by Optical Flow}

The performance of the optical flow estimation network directly affects the shared semantics. In Table~\ref{table_warp} we compare the semantic segmentation results of CAM-60 branch with or without the optical flow warping in the semantics sharing module. Note that we have also skipped the feature fusion modules in the evaluation. The results depict that with only warping by perspective transformation (P.T.), the segmentation results are relatively poor especially for those classes of small objects, which means the semantics are badly propagated. After applying the warping with the optical flow, the performance has a significant enhancement (11.0$\%$) suggesting the importance of accurate remapping.

\begin{table}[!t]
	\renewcommand{\arraystretch}{1.3}
	\vspace{10pt}
	\caption{Comparisons of optical estimation results on different datasets with different network structures.}
	\label{table_net_structure}
	\centering
	\resizebox{0.38\textwidth}{!}{
	\begin{tabular}{|c|c|c|c|c|}
		\hline
		\multirow{3}{*}{\textbf{Network Structure}} & \multirow{2}{*}{\textbf{AEPE}} & \multicolumn{3}{c|}{\textbf{Unsupervised Loss}} \\ 
		\cline{3-5}
		&  & $L_1$ & SSIM & Smooth \\
		\cline{2-5}
		& Data\_Sim & \multicolumn{3}{c|}{Data\_Real} \\
		\hline
		Loosely-coupled & \textbf{1.86} & \textbf{15.4} & \textbf{0.376} & \textbf{0.018} \\
		\hline
		Tightly-coupled & 3.67 & 17.5 & 0.422 & 0.024 \\
		\hline
	\end{tabular}}
\end{table}

\begin{table}[!t]
	\renewcommand{\arraystretch}{1.3}
	\caption{Comparisons of optical estimation results on different datasets with different training schedules.}
	\label{table_train_schedule}
	\centering
	\resizebox{0.40\textwidth}{!}{
	\begin{tabular}{|c|c|c|c|c|c|}
		\hline
		\multirow{3}{*}{\textbf{Training Schedule}} & \multirow{2}{*}{\textbf{AEPE}} & \multicolumn{3}{c|}{\textbf{Unsupervised Loss}} \\ 
		\cline{3-5}
		&  & $L_1$ & SSIM & Smooth \\
		\cline{2-5}
		& Data\_Sim & \multicolumn{3}{c|}{Data\_Real} \\
		\hline
		Chairs & 1.07 & 22.8 & 0.429 & 0.192 \\
		\hline
		Chairs-Simulate	& \textbf{0.08} & 18.7 & 0.421 & 0.030 \\
		\hline
		Chairs-Simulate-Real & 1.86 & \textbf{15.4} & \textbf{0.376} & \textbf{0.018} \\
		\hline
	\end{tabular}}
\end{table}

\begin{table}[!t]
	\renewcommand{\arraystretch}{1.3}
	\caption{Comparisons of semantic segmentation results on Data\_Real w/ or w/o warping by optical flow.}
	\label{table_warp}
	\centering
	\resizebox{0.45\textwidth}{!}{
	\begin{tabular}{|c|c|c|c|c|c|c|c|}
		\hline
		\multirow{2}{*}{\textbf{Warping}} &  \multicolumn{7}{c|}{\textbf{Mean IoU of Semantic Segmentation for CAM-60 ($\%$)}} \\ 
		\cline{2-8}
		& BG & Road	& Person & Car & Barrier & Cycle & Avg. \\
		\hline
		P.T. & 98.0 & 96.5 & 33.9 & 84.6 & 35.7 & 50.0 & 66.4 \\
		\hline
		P.T. + Flow & \textbf{98.9} & \textbf{98.5} & \textbf{61.6} & \textbf{92.4} & \textbf{53.0} & \textbf{59.9} & \textbf{77.4} \\
		\hline
	\end{tabular}}
\end{table}

\begin{table}[!t]
	\renewcommand{\arraystretch}{1.3}
	\caption{Comparisons of semantic segmentation results on Data\_Real dataset with different feature fusion modules.}
	\label{table_ffm}
	\centering
	\resizebox{0.45\textwidth}{!}{
	\begin{tabular}{|c|c|c|c|c|c|c|c|}
		\hline
		\multirow{3}{*}{\makecell*[c]{\textbf{Feature} \\ \textbf{Fusion} \\ \textbf{Module}}} &  \multicolumn{7}{c|}{\multirow{2}{*}{\textbf{Mean IoU of Semantic Segmentation for CAM-60 ($\%$)}}} \\ 
		& \multicolumn{7}{c|}{~} \\
		\cline{2-8}
		& BG & Road	& Person & Car & Barrier & Cycle & Avg. \\
		\hline
		None & 98.9 & 98.5 & 61.6 & 92.4 & 53.0 & 59.9 & 77.4 \\
		\hline
		Basic & 99.1 & 98.8 & 69.2 & 93.6 & 58.1 & 62.5 & 80.2 \\
		\hline
		Residual & \textbf{99.3} & \textbf{99.0} & 71.1 & \textbf{94.2} & \textbf{64.4} & 63.1 & 81.8 \\
		\hline
		Bottleneck & 99.2 & \textbf{99.0} & \textbf{71.7} & 94.1 & 62.0 & \textbf{65.5} & \textbf{81.9} \\
		\hline
	\end{tabular}}
	\vspace{-0.15in}
\end{table}

\subsubsection{Semantic Feature Fusion with Different Types of Blocks}
\label{exp_compare_ffm}

Table~\ref{table_ffm} illustrates the comparisons of integrating different types of feature fusion blocks in CAM-60 branch as an example. We can find that even the simplest basic block can dramatically boost the final segmentation performance. The bottleneck type achieves similar outputs to the residual type in most classes as well as the total average, although it has much less parameters and needs lower computation.

\section{Conclusions}

In this letter we demonstrate how to boost the performance of a scene parsing task for real-time autonomous driving applications with shared semantics. A semantics sharing and fusion framework was proposed to propagate semantic features between two cameras with different perspectives and overlapped views. The shared semantics can not only reduce the duplicable computation in feature extraction procedures, but also refine the segmentation results of both cameras. In the future work we will further investigate to sharing semantics in video scene parsing to realize a more compact and faster semantic perception system.

\addtolength{\textheight}{0cm}   




%
%
%


\bibliographystyle{IEEEtran}
\bibliography{IEEEabrv,ref}

\begin{thebibliography}{10}
\providecommand{\url}[1]{#1}
\csname url@samestyle\endcsname
\providecommand{\newblock}{\relax}
\providecommand{\bibinfo}[2]{#2}
\providecommand{\BIBentrySTDinterwordspacing}{\spaceskip=0pt\relax}
\providecommand{\BIBentryALTinterwordstretchfactor}{4}
\providecommand{\BIBentryALTinterwordspacing}{\spaceskip=\fontdimen2\font plus
\BIBentryALTinterwordstretchfactor\fontdimen3\font minus
  \fontdimen4\font\relax}
\providecommand{\BIBforeignlanguage}[2]{{%
\expandafter\ifx\csname l@#1\endcsname\relax
\typeout{** WARNING: IEEEtran.bst: No hyphenation pattern has been}%
\typeout{** loaded for the language `#1'. Using the pattern for}%
\typeout{** the default language instead.}%
\else
\language=\csname l@#1\endcsname
\fi
#2}}
\providecommand{\BIBdecl}{\relax}
\BIBdecl

\bibitem{siam2018comparative}
M.~Siam \emph{et~al.}, ``A comparative study of real-time semantic segmentation
  for autonomous driving,'' in \emph{Proc. {IEEE} Conf. Comput. Vision Pattern
  Recognit. Workshops}, 2018, pp. 587--597.

\bibitem{tesla}
\BIBentryALTinterwordspacing
Tesla, ``Tesla autopilot.'' [Online]. Available:
  \url{https://www.tesla.com/autopilot}
\BIBentrySTDinterwordspacing

\bibitem{long_fully_2015}
J.~Long \emph{et~al.}, ``Fully convolutional networks for semantic
  segmentation,'' in \emph{Proc. {IEEE} Conf. Comput. Vision Pattern
  Recognit.}, 2015, pp. 3431--3440.

\bibitem{badrinarayanan_segnet:_2017}
V.~Badrinarayanan \emph{et~al.}, ``Segnet: {A} deep convolutional
  encoder-decoder architecture for image segmentation,'' \emph{{IEEE} Trans.
  Pattern Anal. Mach. Intell.}, vol.~39, no.~12, pp. 2481--2495, 2017.

\bibitem{zhao2017pyramid}
H.~Zhao \emph{et~al.}, ``Pyramid scene parsing network,'' in \emph{Proc. {IEEE}
  Conf. Comput. Vision Pattern Recognit.}, 2017, pp. 2881--2890.

\bibitem{chen2018encoder}
L.-C. Chen \emph{et~al.}, ``Encoder-decoder with atrous separable convolution
  for semantic image segmentation,'' in \emph{Proc. Euro. Conf. Comput. Vis.},
  2018.

\bibitem{yu2018bisenet}
C.~Yu \emph{et~al.}, ``Bisenet: Bilateral segmentation network for real-time
  semantic segmentation,'' in \emph{Proc. Euro. Conf. Comput. Vis.}, 2018, pp.
  325--341.

\bibitem{zhao2018icnet}
H.~Zhao \emph{et~al.}, ``Icnet for real-time semantic segmentation on
  high-resolution images,'' in \emph{Proc. Euro. Conf. Comput. Vis.}, 2018, pp.
  405--420.

\bibitem{howard2019searching}
A.~Howard \emph{et~al.}, ``Searching for mobilenetv3,'' in \emph{Proc. {IEEE}
  Int. Conf. Comput. Vis.}, 2019, pp. 1314--1324.

\bibitem{yin2018geonet}
Z.~Yin \emph{et~al.}, ``Geonet: Unsupervised learning of dense depth, optical
  flow and camera pose,'' in \emph{Proc. {IEEE} Conf. Comput. Vision Pattern
  Recognit.}, 2018, pp. 1983--1992.

\bibitem{mayer2016a}
N.~Mayer \emph{et~al.}, ``A large dataset to train convolutional networks for
  disparity, optical flow, and scene flow estimation,'' in \emph{Proc. {IEEE}
  Conf. Comput. Vision Pattern Recognit.}, 2016, pp. 4040--4048.

\bibitem{godard2017unsupervised}
C.~Godard \emph{et~al.}, ``Unsupervised monocular depth estimation with
  left-right consistency,'' in \emph{Proc. {IEEE} Conf. Comput. Vision Pattern
  Recognit.}, 2017, pp. 270--279.

\bibitem{dosovitskiy2015flownet}
A.~Dosovitskiy \emph{et~al.}, ``Flownet: Learning optical flow with
  convolutional networks,'' in \emph{Proc. {IEEE} Conf. Comput. Vision Pattern
  Recognit.}, 2015, pp. 2758--2766.

\bibitem{hui2018liteflownet}
T.-W. Hui \emph{et~al.}, ``Liteflownet: A lightweight convolutional neural
  network for optical flow estimation,'' in \emph{Proc. {IEEE} Conf. Comput.
  Vision Pattern Recognit.}, 2018, pp. 8981--8989.

\bibitem{sun2018pwc}
D.~Sun \emph{et~al.}, ``Pwc-net: Cnns for optical flow using pyramid, warping,
  and cost volume,'' in \emph{Proc. {IEEE} Conf. Comput. Vision Pattern
  Recognit.}, 2018, pp. 8934--8943.

\bibitem{xu2018dynamic}
Y.-S. Xu \emph{et~al.}, ``Dynamic video segmentation network,'' in \emph{Proc.
  {IEEE} Conf. Comput. Vision Pattern Recognit.}, 2018, pp. 6556--6565.

\bibitem{zhu2019improving}
Y.~Zhu \emph{et~al.}, ``Improving semantic segmentation via video propagation
  and label relaxation,'' in \emph{Proc. {IEEE} Conf. Comput. Vision Pattern
  Recognit.}, 2019, pp. 8856--8865.

\bibitem{jin2017video}
X.~Jin \emph{et~al.}, ``Video scene parsing with predictive feature learning,''
  in \emph{Proc. {IEEE} Int. Conf. Comput. Vis.}, 2017, pp. 5580--5588.

\bibitem{li2018low}
Y.~Li \emph{et~al.}, ``Low-latency video semantic segmentation,'' in
  \emph{Proc. {IEEE} Conf. Comput. Vision Pattern Recognit.}, 2018, pp.
  5997--6005.

\bibitem{wang2019learning}
X.~Wang \emph{et~al.}, ``Learning correspondence from the cycle-consistency of
  time,'' in \emph{Proc. {IEEE} Conf. Comput. Vision Pattern Recognit.}, 2019,
  pp. 2566--2576.

\bibitem{lee2019sfnet}
J.~Lee \emph{et~al.}, ``Sfnet: Learning object-aware semantic correspondence,''
  in \emph{Proc. {IEEE} Conf. Comput. Vision Pattern Recognit.}, 2019, pp.
  2278--2287.

\bibitem{li2017foveanet}
X.~Li \emph{et~al.}, ``Foveanet: Perspective-aware urban scene parsing,'' in
  \emph{Proc. {IEEE} Int. Conf. Comput. Vis.}, 2017, pp. 784--792.

\bibitem{jiao2019geometry}
J.~Jiao \emph{et~al.}, ``Geometry-aware distillation for indoor semantic
  segmentation,'' in \emph{Proc. {IEEE} Conf. Comput. Vision Pattern
  Recognit.}, 2019, pp. 2869--2878.

\bibitem{hoyer2019short}
L.~Hoyer \emph{et~al.}, ``Short-term prediction and multi-camera fusion on
  semantic grids,'' in \emph{Proc. {IEEE} Int. Conf. Comput. Vis. Workshops},
  2019.

\bibitem{szeliski2010computer}
R.~Szeliski, \emph{Computer vision: algorithms and applications}.\hskip 1em
  plus 0.5em minus 0.4em\relax Springer Science \& Business Media, 2010.

\bibitem{wang2004image}
Z.~Wang \emph{et~al.}, ``Image quality assessment: from error visibility to
  structural similarity,'' \emph{{IEEE} Trans. Image Process.}, vol.~13, no.~4,
  pp. 600--612, 2004.

\bibitem{heise2013pm}
P.~Heise \emph{et~al.}, ``Pm-huber: Patchmatch with huber regularization for
  stereo matching,'' in \emph{Proc. {IEEE} Int. Conf. Comput. Vis.}, 2013, pp.
  2360--2367.

\end{thebibliography}

\end{document}